# Language-Specific Representation of Emotion-Concept Knowledge Causally Supports Emotion Inference

Ming Li,[1,2,#] Yusheng Su,[3,#] Hsiu-Yuan Huang,[4] Jiali Cheng,[5] Xin Hu,[1,2,6] Xinmiao Zhang,[1,2] Huadong Wang,[3] Yujia Qin,[3] Xiaozhi Wang,[3] Kristen A. Lindquist,[7] Zhiyuan Liu,[3,**] Dan Zhang[1,2,*]

[1]Department of Psychology, Tsinghua University

[2]Tsinghua Laboratory of Brain and Intelligence, Tsinghua University

[3]Department of Computer Science and Technology, Tsinghua University

[4]School of Computer and Communication Engineering, University of Science and Technology Beijing

[5]Miner School of Computer and Information Sciences, University of Massachusetts Lowell

[6]Department of Psychiatry, University of Pittsburgh

[7]Department of Psychology and Neuroscience, University of North Carolina

## Contact Information

Correspondence concerning this article should be addressed to Dan Zhang (dzhang@tsinghua.edu.cn) and Zhiyuan Liu (liuzy@tsinghua.edu.cn).

[#]These authors contributed equally.

**Summary**

Humans no doubt use language to communicate about their emotional experiences, but does language in turn help humans understand emotions, or is language just a vehicle of communication? This study used a form of artificial intelligence (AI) known as large language models (LLMs) to assess whether language-based representations of emotion causally contribute to the AI's ability to generate inferences about the emotional meaning of novel situations. Fourteen attributes of human emotion concept representation were found to be represented by the LLM's distinct artificial neuron populations. By manipulating these attribute-related neurons, we in turn demonstrated the role of emotion concept knowledge in generative emotion inference. The attribute-specific performance deterioration was related to the importance of different attributes in human mental space. Our findings provide a proof-in-concept that even a LLM can learn about emotions in the absence of sensory-motor representations and highlight the contribution of language-derived emotion-concept knowledge for emotion inference.

**Introduction**

At the end of William Shakespeare's Hamlet, Horatio looks at Hamlet and says, 'Now cracks a noble heart. Good night, sweet prince, / And flights of angels sing thee to thy rest.' Although Horatio's facial-bodily expression and tone of voice are imperceptible from this line, the reader can still infer Horatio's grief and admiration. Humans have long been accustomed to communicating individual mental experiences through language (Baumard et al., 2022; Lindquist et al., 2022). However, language's role in inferring others' emotions remains a matter of debate.

This question implicates an ongoing debate regarding the relationship between language and discrete emotion distinctions (Satpute & Lindquist, 2021). Among various perspectives, the traditional ones hold that language processing has no or little effect on the emotion experienced in a given situation because human emotions are considered as categories "embodied" in concrete sensory-motor experiences generated by discrete emotional mechanisms within the brain and body (Ekman & Cordaro, 2011; Tracy & Randles, 2011). In this view, language primarily plays a communicative role in conveying the meaning of an emotional experience after the fact (Ekman, 1992; Keltner, Dacher, Cordaro, 2015).

In contrast, the constructivist account (Barrett, 2017; Lindquist, 2013) suggests that language experience is involved in shaping the conceptual boundaries of discrete emotions, thereby allowing language-derived knowledge to play an essential role in dictating which emotions are experienced in which contexts. In this view, emotion concepts represent and help differentiate the abstract feature space that comprises emotional experiences. As individuals acquire emotion concepts during early experience, these concepts prospectively warp future experiences of feature space. Similar considerations on color cognition, for illustration, show that congenitally blind people understand color space reasonably well (i.e., which colors are more vs. less similar to one another) solely as a product of learning from other human's linguistic descriptions of the sensorimotor features of color space (Bi, 2021; Shepard & Cooper, 1992). Suppose language also plays a constitutive role in emotion conceptualization. In that case, human language should be a sufficient, though not the only, way to learn and infer the rich meaning of discrete emotions (Satpute & Lindquist, 2019).

Developmental research supports this perspective by showing that language experience, as a source of emotion-concept knowledge (Nook et al., 2017; Shablack et al., 2020), may contribute to children's development of increasingly discrete emotional experiences (Hoemann et al., 2020; Nencheva et al., 2023; Streubel et al., 2020). Increasing behavioral studies further revealed that individual (Brooks & Freeman, 2018; Hu et al., 2021) and cultural (Barrett et al., 2019; Gendron et al., 2014) differences in emotion-concept knowledge correspond to differences in the sensorimotor representations of emotional facial behaviors. These findings provide correlational evidence for the functional importance of language in emotion conceptualization.

Despite the promising progress, causal evidence for the role of language-grounded knowledge in emotion inference is currently lacking (Lindquist, 2017). To this end, researchers have explored manipulations of concept-knowledge representations in the human mind. Through priming techniques (Maxfield, 1997), studies have demonstrated that presenting emotion concept words (e.g., "anger") before behavioral tasks requiring participants to draw emotional inferences about the meaning of facial behaviors shapes their subsequent perceptions (Gendron et al., 2012; Lindquist et al., 2015; Nook et al., 2015). The mechanism of priming is to influence the human mind's access to emotion concepts by pre-activating relevant knowledge representations using priming cues. However, manipulating the access does not directly manipulate knowledge representation *per se* (Firestone & Scholl, 2014, 2015), leaving the evidence circumstantial.

A more direct approach involves investigating the behavioral consequences of neurological disorders (Jastorff et al., 2016; Lindquist et al., 2014) or stimulation (Long et al., 2023) in brain regions potentially associated with abstract concept-knowledge representation, i.e., semantic memory (Tulving, 1972). For instance, patients with semantic dementia, characterized by lesions in the anterior temporal lobe (ATL), the "hub" of semantic memory (Patterson et al., 2007), are unable to make inferences about the discrete categorical meaning of facial behaviors (Lindquist et al., 2014). However, the ATL is but one brain region involved in the representation of emotion concepts and is also functionally connected with areas responsible for high-order sensory processing (Mahon & Caramazza, 2008; Sabsevitz et al., 2005), which may support the integration of sensory-grounded modality-specific information (Bi, 2021; Popham et al., 2021). Due to the limited understanding of ATL and the difficulty of excluding sensory experience, the extent to which emotion inference relies on non-sensory language-derived knowledge remains uncertain.

Considering the divergence between different theoretical claims, the controversy may be alternatively formulated as whether, without access to sensory-motor representations, discrete emotion concepts can still be learned and inferred through language. Recent developments in large language models (LLMs; Brown et al., 2020; Petroni et al., 2019) provide a valuable opportunity to verify this hypothesis by showing the purely language-based representation of emotion-concept knowledge and its causal support for emotion inference.

The investigation of language-derived knowledge in LLMs rests on the assumption that semantically relevant attributes of concepts are reflected in the pattern of linguistic symbol use (Grand et al., 2022; Günther et al., 2019). Thus, text-based computing offers the opportunity to mine linguistic-cultural phenomena for their psychological meanings (Jackson et al., 2022; Michel et al., 2011). Many nascent studies have shown that, in the absence of human annotation, LLMs can learn various domains of human knowledge, exclusively on the basis of human language use (Aspillaga et al., 2021; Ettinger, 2020; Schramowski et al., 2022). For instance, recent evidence shows that LLMs can generate moral judgments akin to those made by humans, solely based on learning of human language use (Dillion et al., 2023).

LLMs are not only useful as concept proofs about the ability of AI to represent complex aspects of human psychology based on human language use alone. LLMs can also be used experimentally to understand how knowledge is applied to novel judgments. The knowledge obtained from pre-training is stored in LLMs and can be selectively activated for different downstream tasks. In this form, the pre-trained LLMs may serve as ideal "subjects" who learn about the world exclusively through human language, and we can then inspect their use of emotion-concept knowledge in tasks where the LLM must make an inference about the meaning of a novel emotional situation.

Critically, LLMs can be manipulated more easily than the human brain to understand the relationship between language-specific knowledge representation and emotion inference. Specifically, the artificial neurons in LLMs can be selectively manipulated for their functional relevance to specific concept attributes. In turn, the role of neurons associated with certain content in LLMs can then be causally assessed vis a vis their role in facilitating the LLM's performance on inference tasks (Wang et al., 2022). This practice resembles the neural stimulation techniques used in neuroscience research (Hallett, 2007;

Walsh & Cowey, 2000) but with a higher level of precision (e.g., at the level of single artificial neuron) than is available in human subjects.

Considering the computational principles of language processing shared by LLMs and humans (Goldstein et al., 2022), findings from LLMs have the potential to shed light on the language-based mechanism underlying human emotion inference. This stance of analyzing computational models from a human-like perspective (Grand et al., 2022; Zhou et al., 2022) has been recognized as beneficial to explore the functional emergence of human cognitive abilities and to address questions that are difficult to answer through human studies alone (Dillion et al., 2023; Doerig et al., 2023; Frank, 2023; Kanwisher et al., 2023).

In the present study, we aimed to explore the language-specific representation of emotion-concept knowledge and its support for emotion inference by LLMs (Fig. 1). To justify this possibility, the model we chose needed to be mature enough to perform emotion-related tasks while untuned by user feedback, and also provide open parameters for manipulation experiments. We utilized RoBERTa (Y. Liu et al., 2019) in our experiments as the base model, which is a typical LLM that is pre-trained by filling in randomly masked parts of a massive corpus. Since the goal of pre-training is to reproduce as much as possible real human language use, we hypothesize that this LLM can learn human emotion-concept knowledge from the pre-training corpus.

Then the LLM was instructed to perform 27 emotion inference tasks, each requiring it to infer a specific emotion from the same task dataset (Methods). To better elicit the LLM's performance, instead of manually designing questioning templates (e.g., "Does the following sentence express the emotion of grief:"), we trained 27 emotion-specific task prompts using the training set, i.e., prompt-tuning (P. Liu et al., 2021). For example, to instruct the LLM to infer remorse, the input would be a combination of the remorse-specific prompt (an optimized questioning template), the text to be inferred, and a placeholder that accepts "yes" or "no" as the output. The LLM's parameters are frozen during training the task prompts, and then both LLM's parameters and task prompts are immutable for testing.

Since the task prompts determine which emotion the LLM infers from the same task dataset, the LLM's neuron activations in respond to emotion-specific prompts were expected to represent knowledge of corresponding emotion concepts. To validate the representational content of LLM-based emotion-concept knowledge, we conducted behavioral experiments to obtain human ratings on 14 attributes of emotion concept from existing literature, including core affects, prototypical expressions, and antecedent appraisals. We compared these ratings with the LLM's representations from a higher-level functional perspective through representational similarity analysis (RSA) (Kriegeskorte et al., 2008). Subsequently, guided by the representational space of the human ratings, we could locate and manipulate the artificial neurons in the LLM relevant to a conceptual attribute (e.g., valence) of emotions to investigate their causal contributions for emotion inference tasks. We further explored the association between the language-based contribution of different conceptual attributes and their importance in human mental space, which could provide evidence for a deeper understanding of the constitution and inference of emotion concepts.

## Results

### LLM Infers Emotions Based on Shared Conceptualization

After optimizing 27 emotion-specific prompts to infer the corresponding emotion on the training set, we reported the average accuracies of task prompts over different random seeds by evaluating the test set (Table S1, see also Fig. 3a). The average accuracy for each of the 27 emotion inference tasks varied from 68.04% (realization) to 96.43% (gratitude).

The LLM's inference accuracy was positively related to rater agreement on dataset annotations. Pearson's $r(25)$ was .797 with $p < .001$, suggesting that the more agreement human raters had on a particular type of emotional scenarios, the more accurate the LLM's inference is (see also Table S1 and Fig. S1).

### Language-Specific Representations of Emotion-Concept Knowledge

As the LLM's neuron activations in respond to an emotion-specific prompt solely without text samples were considered to represent the corresponding emotion concepts, we investigated their representational content using RSA. This technique aims to evaluate the second-order similarity between the representational dissimilarity matrices (RDMs) of every single artificial neuron's activation and the RDMs of 14 conceptual attributes of emotion, which were obtained from human rating experiments (Methods). See Figure 2b for illustration.

The second-order similarities exhibited that each of the 14 conceptual attributes was significantly related to a subset of artificial neurons, which were distributed in all layers of the LLM rather than concentrated in specific layers. The results are shown in Figures S5 and S6, FDR corrected $q < .01$, one-tailed sign-rank test.

According to the rank of relatedness (Kendall's *tau*) between artificial neurons and conceptual attributes, the most relevant neurons for different attributes were less overlapping. The example of the top 4,000 attribute-specific neurons is shown in Figure 2c; see Figure S7 for results of different top N.

### Emotion-Concept Knowledge Causally Contributes to Emotion Inference

The reliance of emotion inference on language-derived emotion-concept knowledge was then revealed by manipulating attribute-specific neurons while inferring 27 emotions (Fig. 2d). See Methods for the details of the manipulation experiment. Compared to the original accuracy without manipulation, we found a drop in the accuracy of emotion inference on LLM with selective manipulation (Fig. 3a and Fig. S8).

This deterioration of inference performance still held when compared to randomly manipulating the same number of neurons, suggesting the unique causal contribution of emotion-concept knowledge representations *per se* (Fig. S9). Significance was determined by a one-tailed paired *t*-test on accuracy drop averaged across random seeds and conceptual attributes, FDR corrected $p < .05$. The most prominent performance deterioration arose when manipulating the top 4,000 attribute-specific neurons, shown in Figure 3b.

The possible differences in the knowledge contribution for inferring different emotions were examined by testing the heterogeneity of 27 tasks' performance deterioration. Hartigan's dip test showed no evidence of significant heterogeneity regarding the reliance on emotion-concept knowledge for different tasks, with minimum *p* = .127 for every conceptual attribute and every number of manipulated neurons (Table S2).

**Importance in Mental Space Predicts Language-Based Contribution of Different Attributes**

To further explain the contributions of different conceptual attributes to emotion inference (Fig. 4a), we also estimated the weight of different knowledge in the human mental representation of emotion concepts (Fig. 4b) and compared the evidence from two parts. See Methods for the details of human experiment and related analysis.

For the LLM with access only to natural language, 12 of the 14 conceptual attributes contribute significantly to emotion inference under specific numbers of manipulated neurons in varying degrees, except for "self-relativity" and "disgust-face" (Fig. 4a; for details, see Fig. S10). Significance was determined by a one-tailed paired *t*-test on accuracy drop averaged across random seeds and inference tasks, FDR corrected *p* < .05.

For humans with both normal language and sensory functions, their mental representations reflect 12 of the 14 conceptual attributes, except for "arousal" and "other-relativity". Two-tailed signed-rank test with bootstrap sampling, FDR corrected *p* < .001 across all attributes (Fig. 4b).

We further demonstrated that the language-based knowledge contribution to emotion inference was not independent, but significantly related to knowledge weight in human mental representation. The strongest correlation arose when manipulating the top 4,000 attribute-specific neurons in the LLM, with Fisher-based average *r* = .473, *t*(26) = -16.714, *SD* = .030, *p* < $10^{-15}$ (Fig. 4c). In addition, as the number of manipulated neurons increased from 1,500 to 6,000, an inverted-U shaped relationship was observed (Fig. 4d), suggesting that there may be "floor effects" and "ceiling effects" (Lim et al., 2015) to manifest the reliance of emotion inference on emotion-concept knowledge in the LLM.

## Discussion

In the present study, we adopted a human-like perspective to investigate the representation of emotion-concept knowledge in a LLM in response to the theoretical debates about the relationship between language and discrete emotion distinctions. Our results provide a proof-in-concept that, even without any access to sensory-motor representations, various attributes of emotion concepts can still be abstracted from language experience and causally support emotion inferences. This computational evidence makes no demand for the experiential feelings once thought essential for discrete emotion differentiation, though, it correlated well with behavioral evidence of the importance of these conceptual attributes.

The findings that the LLM's neuron activations effectively represented emotion-concept knowledge extend our understanding of language-accessible human knowledge to the emotion domain (Fig. S5). While previous studies have preliminarily shown the association between language learning and emotion conceptualization (Hoemann et al., 2021; Shablack et al., 2020), we further demonstrated that attributes of emotion concept could derive from the sensory-independent language experience, i.e., the statistical regularities among the linguistic symbols. This computational evidence supports a valuable but understudied hypothesis: humans can learn emotion concepts directly from everyday language use (Hoemann et al., 2019, 2020).

The consistency of LLM's emotion inference performance with human raters' agreement on emotion annotations further indicates that what is reflected in the large-scale language corpus is a shared understanding among individuals (Table S1 and Fig. S1). Intriguingly, the understanding includes not only cognitive appraisal attributes, but also more experiential attributes, i.e., the associated core affects and similarity with prototypical facial expressions (Supplemental Method and Fig. S2). Given the computational nature of LLMs (Blank, 2023), it would be interesting to elucidate further how individual differences in abstract symbol use influence emotional development (Camacho et al., 2023; Nook et al., 2017).

Moreover, the LLM's knowledge-related artificial neurons were distributed across all layers (Fig. S6), possibly implying the involvement of both low-level (e.g., phrasal features) and high-level (e.g., syntactic and semantic features) linguistic regularities for emotion conceptualization (Jawahar et al., 2020). The distributions of the artificial neurons corresponding to different conceptual attributes also tended to be distinct (Figs. 2c and S7), suggesting possibly unique contributions of these attributes. These results may inspire further exploration of the neural mechanisms underlying language-derived knowledge representations about emotion concepts.

Most importantly, our rigorous manipulation of attribute-specific neurons in the LLM (Figs. 2d and 3a-b) examined the functional validity of the language-specific representations of emotion-concept knowledge in emotion inference. This result could contribute to a central and ongoing debate in emotion science about the nature of human emotion categories (Adolphs et al., 2019; Barrett, 2006a; Cowen & Keltner, 2021; Fridlund, 2017), i.e., whether language only superficially conveys emotional experiences after the fact, or whether emotion experiences are constructed in part via abstract conceptual category knowledge acquired via language (Barrett & Westlin, 2021). Whereas previous lesion studies revealed the necessity of semantic memory for discrete emotion differentiation (Lindquist et al., 2014), we used the LLM to demonstrate further that language-specific knowledge representations are sufficient, at least in principle, to differentiate discrete emotions from meaningful contexts (Figs. 3b, 4a; see Figs. S8-10 for whole results). By pointing out the weak heterogeneity in the knowledge contribution to 27 emotions (Table S2), we suggest a unifying mechanism for LLMs and possibly also for the human brain to infer various emotions. Our view is reinforced by the recent neuroimaging findings that one broad ensemble containing multiple brain networks represents a range of emotions (Horikawa et al., 2020) rather than distinct emotions consistently and specifically activating local brain regions (Lindquist et al., 2012).

It is worth noting that the language-based emotion inference mechanism is not exclusive of sensory-motor processes, nor of other semantic processes integrated with sensory-motor experiences, such as meaning-making (Satpute & Lindquist, 2019) and prototype-matching (Binetti et al., 2022). Instead, our comparison between the language-based knowledge contribution and the knowledge weight in human mental space suggests that language has limitations in supporting emotion inference. For example, albeit the conceptual similarity with "disgust-face" can be reflected by both the LLM's artificial neurons (Fig. S5-6; see also Fig. 2c) and human mental representation (Fig. 4b), its language-specific representation *did not* effectively contribute to emotion inference (Fig. S7). Based on this study, it is hard to elucidate whether the language-specific representation of specific conceptual attributes ("self-relativity" and "disgust-face") needs to be combined with other experiential information to be functional. However, the correlation between the language-based knowledge contributions and the knowledge weights in human mental representation (Fig. 2c-d) illustrates that human conceptualization of discrete emotions is inextricably linked to the functionality of the language-derived knowledge.

Since LLMs and the human brain have been suggested to share similar computational principles for language (Goldstein et al., 2022), LLMs can serve as a potential reference in the future to help us understand the language-dependent algorithms that the human brain relies on to infer emotions. Future research could also use LLMs to investigate how different semantic processes (Bi, 2021; Popham et al., 2021) drive the supramodal representation of emotions in the brain. For example, growing evidence suggested that the brain can convergently process and integrate emotional cues across modalities (e.g., facial expressions and prosody) and represent their conceptual meaning in amodal areas (Schirmer & Adolphs, 2017). Suppose the activity of these amodal areas during emotion perception fits with the LLM's hidden state values. In that case, possible neural mechanisms of language-dependent semantic processing involved in making emotional meanings from sensory input can be revealed.

In conclusion, recent advances in LLMs provide a rare opportunity to test the independent role of language in conceptualizing and inferring discrete emotions. The fact that even an LLM can learn the rich meaning of emotions solely through human language reminds researchers in the field of emotion not to neglect the possible effects of language constructions when interpreting their results. How two basic functions, language and sensory-motor, interact with each other in emotion-related cognitive and neural mechanisms should be the focus of future research.

**Limitations of the Study**

Since our aim was to investigate whether emotion concepts derived from human language are sufficient to help differentiate emotional meaning from novel scenarios, the use of the LLM completely precluded access to sensory-motor representations. In this context, our results cannot answer, and are not intended to answer, whether emotional feelings are rooted in language use. Also, to conceptually demonstrate the possibility that language has an independent role in discrete emotion distinctions, we chose an LLM that balances performance, stability, and openness as the subject of our study. This does not promise that LLMs trained with other strategies will give evidence of the same strength.

**Author Contributions**

Conceptualization: D. Z. and Z. L.; Methodology: Y. S., H. W., Y. Q., and X. W.; Formal Analysis: M. L., Y. S., X. H., and X. Z.; Investigation: M. L., H. H., and J. C.; Writing – Original Draft Preparation: M. L., D. Z., K. A. L., and Y. S.; Writing – Review & Editing: X. H., X. Z., Z. L., H. H., J. C., H. W., Y. Q., and X. W.; Funding Acquisition: D. Z. and Z. L.; Project Administration: M. L. and Y. S.; Resources: H. W., Y. Q., and X. W.; Software: M. L., Y. S., H. H., and J. C.; Supervision: D. Z.; Visualization: M. L.

**Acknowledgments**

This work was supported by the National Natural Science Foundation of China (61977041, 62236004), the National Key Research and Development Program of China (No. 2020AAA0106500), and the National Science Foundation of China (NSFC) and the German Research Foundation (DFG) in project Cross-modal Learning (NSFC 62061136001/DFG TRR-169/C1, C4).

**Declaration of interests**

The authors declare no competing interests.

**Data and Code Availability**

All the materials used in this research are widely available or are fully described. The data and the code relevant to the results are publicly available at https://github.com/thunlp/Model_Emotion.

**Ethics Approval**

The Institutional Review Board at the Department of Psychology, Tsinghua University, approved the experimental procedures.

**Informed Consent**

All human participants gave their informed consent.

**Methods**

**Language-Specific Representation of Emotion-Concept Knowledge**

*Emotion Inference Tasks*

The text dataset we employed for emotion inference tasks in this study is GoEmotions (*https://github.com/google-research/google-research/tree/master/goemotions*) (Demszky et al., 2020), which contains 58,009 English Reddit comments, manually labeled with 27 emotions by 82 unique raters. Subsequently, for each of the 27 emotions, this same dataset could be viewed as a different task to infer whether the samples belong to the corresponding emotion category.

Since the same dataset was used for different inference tasks, we used 27 emotion-specific task prompts to instruct the LLM on which emotion should be inferred. For example, to instruct the LLM to infer remorse, the input needs to contain three parts simultaneously. The first part is a text sample taken from the test set representing an emotional scenario that the LLM has not seen before. The second part is a task prompt that instructs the LLM to determine whether the sample expresses remorse or not, which is trained using the training set and the corresponding remorse labels. The last part is a special placeholder ([MASK]) that will be filled in by the LLM with "yes" or "no" as its answer.

Following the authors of the dataset, we divided the dataset into training (80%), development (10%), and test (10%) sets. We did not use the development set in any subsequent operation because its proposed purpose was incompatible with this study.

*Training of Emotion-Specific Task Prompts*

To find valid prompts for each task, we optimized only the prompts (some pseudo tokens that function as questioning templates) while freezing all parameters of the LLM to generate the appropriate answer (yes/no) for the input text (Fig. 2a). Here, freezing parameters of the LLM prevented task information from changing the LLM's learned knowledge. The function of the prompts is to instruct the LLM to perform corresponding tasks, but their format need neither contain emotion words nor be human-readable. In other words, the task prompts and the output answers are related to specific emotion concepts, but do not use their lexical forms.

Considering the reasonable but profound individual differences in emotion-concept knowledge (Binetti et al., 2022; Hu et al., 2021), for each sample, we only considered emotions that most raters agreed to infer, which would reflect a conceptualization shared among raters. To avoid statistical bias, for each emotion inference task, we trained its prompt 12 times with 12 random seeds. The details about prompt training are included in the Supplemental Method.

*Evaluation of LLM's Inference Performance*

All task prompts with all random seeds have been evaluated on the test set, respectively (Table S1). Since the LLM is pre-trained on large-scale human language corpora, it should be more capable of inferring emotions with more shared conceptualization. We then estimated rater agreements for each emotion via Cohen's Kappa (Cohen, 1960), and related them with the LLM's inference performance on the test set (Fig. S1).

*Model-Based Knowledge Representation*

Since these prompts can activate the corresponding LLM task state (Su et al., 2022; Wang et al., 2022), the LLM's neuron activation values in response to the emotion-specific prompts were considered to represent knowledge about the corresponding emotion concept. Hence, we input the trained prompts for each of the 27 emotions without concatenating any text into the LLM to activate the hidden states values, also known as the LLM's artificial neuron activations (Fig. 2b). See Supplemental Method for more details. These activation values of 36,864 artificial neurons were collected for further analysis.

**Validation of Language-Specific Knowledge Representation**

*Human Ratings of Emotion-Concept Knowledge*

To measure the content of emotion-concept knowledge, we chose the most representative attributes of emotion concepts in the existing psychological emotion theories, including core affects (i.e., how emotion might feel; Barrett, 2006b; Russell, 2003), prototypical expressions (i.e., similarities with six stereotypical emotional faces; Du et al., 2014; Levenson, 2011), and antecedent appraisals (i.e., what the antecedents of emotion might be; Clore & Ortony, 2016; Scherer & Fontaine, 2019). Through three independent experiments, we obtained human ratings for 14 conceptual attributes (two core affects, $N = 30$; six prototypical expressions, $N = 30$; six antecedent appraisals, $N = 148$) of 27 emotions from Prolific. All participants were English-speaking. See Supplemental Method and Fig. S2. for more details.

All those conceptual attributes were averaged across repeated ratings as the final attribute scores for each emotion concept. The final scores and their reliabilities are shown in Figs. S3 and S4. In addition to the adopted participants reported above, we excluded 5, 6, and 30 subjects from the three experiments due to failure of the attention check, respectively. The Institutional Review Board at the Department of Psychology, Tsinghua University, approved all experimental procedures. All participants gave their informed consent.

*Searchlight RSA for LLM's Representation of Emotion-Concept Knowledge*

We first built a representational dissimilarity matrix (RDM; Kriegeskorte et al., 2008) for each artificial neuron. An RDM is a symmetric matrix (27 emotions by 27 emotions), where the elements are the Euclidean distances of the random-seed-averaged neuron activation in response to emotion-specific prompts. The RDMs for 14 conceptual attributes were also built, respectively, by calculating the Euclidean distances of emotion concepts' final score on that attribute.

We then conducted the one-tailed sign-rank test to indicate the relatedness between the RDM of each artificial neuron and the RDM of each conceptual attribute, using only the lower triangle of RDMs (Fig. 2b). Due to many calculations, we did not perform the bootstrap method. Instead, we used false discovery rate (FDR) correction (Benjamini & Yekutieli, 2001) to control multiple comparisons for all neurons and all attributes. The significance of Kendall's *tau* values shows the absolute correspondence of the artificial neurons for each conceptual attribute (Fig. S5 and S6), and the rank of Kendall's *tau* values shows the relative correspondences (Figs. 2c and S7).

**Manipulation of Emotion-Concept Knowledge Representation**

*Artificial Neuron Manipulation Experiment*

To examine the potential support of emotion-concept knowledge representation for emotion inference, we input the trained prompts and the test set of scenarios into LLM to infer emotions. During the inference of 27 emotions, we modified the activation values of attribute-specific neurons to zero (Fig. 2d). For each conceptual attribute, the number of manipulated neurons was set uniformly to 1500, 2000, 2500, 3000, 4000, 5000, or 6000. Overall, the selective manipulation operation was repeated 34,020 times, respectively, for 14 conceptual attributes, 27 emotion inference tasks (12 prompts/random seeds per task), and seven levels (the number of manipulated neurons).

To exclude the influence of manipulating neurons *per se*, we randomly select the same number of neurons to manipulate as a control group for every operation. The causal contribution of emotion-concept knowledge was indicated as the difference in accuracy after selective manipulation compared to accuracy after random manipulation, i.e., accuracy drop due to the neuron type (attribute-specific vs. random).

*Task Reliance on Emotion-Concept Knowledge*

For each of the 27 emotion inference tasks at a specific number of manipulated neurons, we estimated its reliance on emotion-concept knowledge by one-tailed paired *t*-test, i.e., accuracy drop averaged across random seeds and conceptual attributes. FDR was corrected to determine the significance of accuracy drops. The example of manipulating 4,000 neurons is shown in Fig. 3.

*Heterogeneity Test for the Reliance of Different Tasks*

To explain whether there are systematic differences in the reliance on emotion-concept knowledge for inferring different emotions, we then tested the heterogeneity of 27 tasks' reliance on specific conceptual attribute with specific number of manipulated neurons. This analysis was conducted by examining the multimodality of random-seed-averaged accuracy drop via Hartigan's dip (Freeman & Dale, 2013; Gu, M; Lai, 1985).

**Explanation of Different Knowledge Contribution**

*Language-Based Knowledge Contribution*

The causal contribution of a specific conceptual attribute to emotion inference was estimated by accuracy drop averaged across random seeds and inference tasks, at a specific number of manipulated neurons. FDR was corrected to determine the significance. The example of manipulating 4,000 neurons is shown in Fig. 4a.

*Concept Similarity Judgment Experiment*

We adopted a similarity judgment task to measure the human mental representation of the 27 emotion concepts (Fig. 4b). Sixty-one English-speaking participants (30 females, mean age = 36 years) were recruited from Prolific and asked to complete the task online. They judged the subjective similarity of 27 emotion concepts (and "neutral" concept) using a 9-point Likert scale (1=most dissimilar, 9=most similar) without criteria cues. These 27 emotion and neutral concepts were presented simultaneously on the screen in word form. However, participants judged the similarity of only the two words with black borders each time. There were no response time limits but instructions to participants to respond by first sense when they hesitated.

We retained similarity scores between 27 emotions (351 pairs) and replaced missing values (3 per participant) with the average score across participants. Then, for each participant, these scores were subtracted by 10 to indicate the dissimilarity (ranging from 1-9) and used to form an individual RDM, i.e., a 27 by 27 symmetric matrix with a diagonal of 0 to indicate that any emotion is equal to itself. Each RDM reflected one participant's mental representation of emotion concepts. The Institutional Review Board at the Department of Psychology, Tsinghua University, approved the experimental procedures. All participants gave their informed consent.

### *RSA for Knowledge Weight in Mental Representation of Emotion Concepts*

To estimate the weight of specific conceptual attributes in the human mental representation of emotion concepts, we conducted RSA to show how well the RDM of a conceptual attribute fit the RDMs of people's emotion-concept representations (Kendall's *tau*; Fig. 4b). The RDM of each of 14 conceptual attributes was related to people's RDMs via the two-tailed signed-rank test, with bootstrap sampling participants and emotions 1,000 times. FDR was corrected to control multiple comparisons across 14 attributes.

### *Comparison between Language-Based Knowledge Contribution and Knowledge Weight in Mental Representation*

The degrees of accuracy drop for 14 conceptual attributes were then correlated with these conceptual attributes' weights in human mental representations, i.e., Kendall's *tau* values. Considering the weak heterogeneity of knowledge contribution across emotion inference tasks (see Table S2), we treated each emotion inference task as a sample set and obtained a series of Pearson's correlation coefficients to determine significance by the one-tailed *t*-test on Fisher's transformed coefficients (for an illustration, see example in Fig. 4c; Silver & Dunlap, 1987).

**Figures**

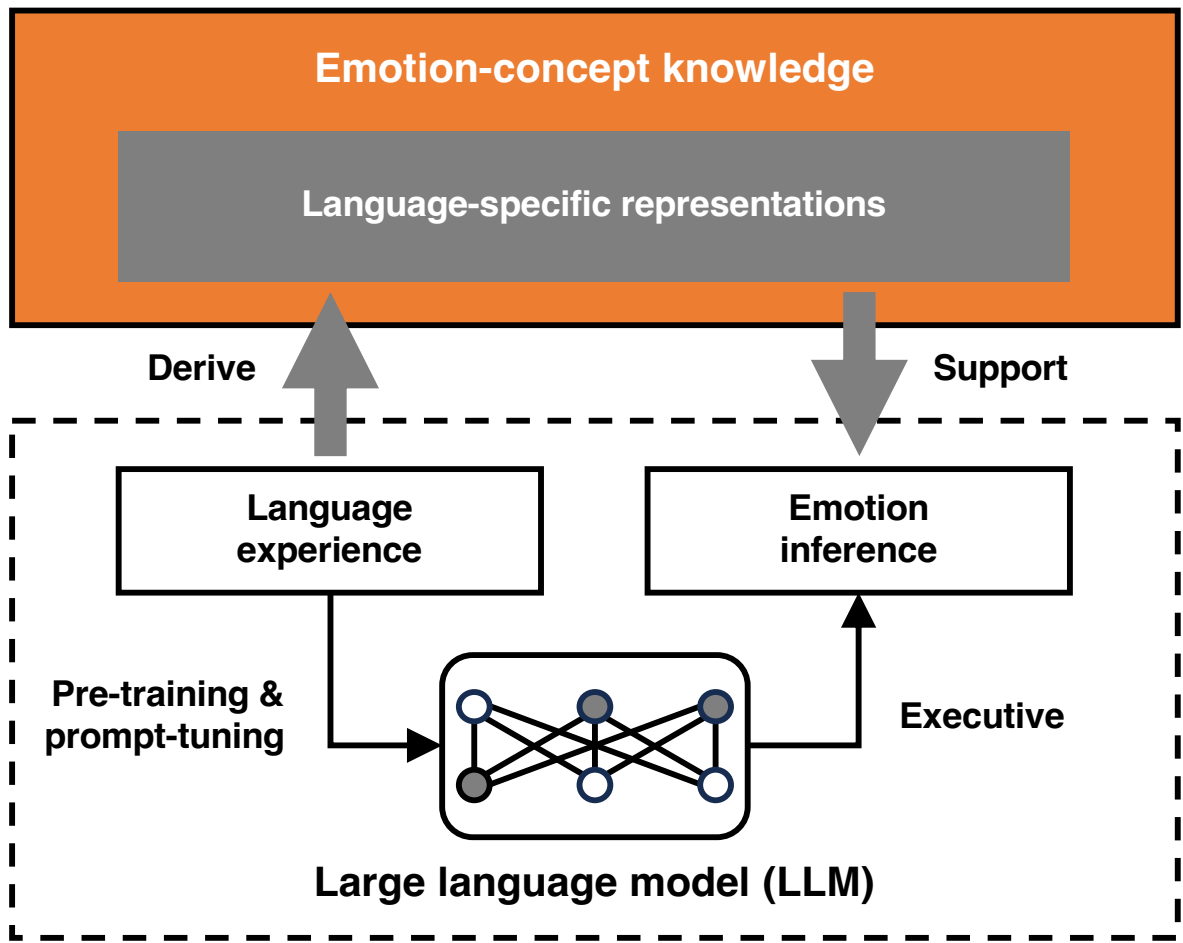

**Figure 1. Computational Approach for Revealing Language-Based Emotional Inference Mechanism.**

To reveal the language-specific representations of emotion-concept knowledge, we took advantage of a pre-trained Large Language Model (LLM). We found that its neuron activations in response to 27 emotion-specific task prompts effectively represent 14 attributes of human emotion concepts. The reliance of emotion understanding on emotion-concept knowledge was then demonstrated by manipulating the activations of attribute-specific artificial neurons in the LLM.

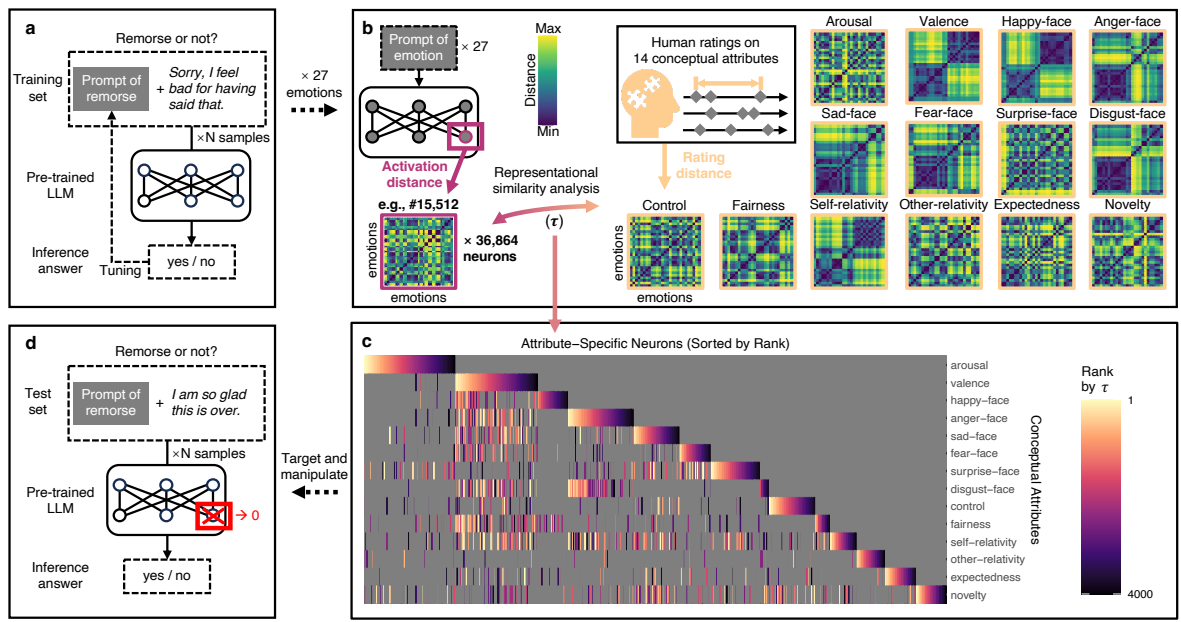

**Figure 2. Procedure of Manipulating the Language-Specific Representations of Emotion-Concept Knowledge during Emotion Inference Tasks.**

Illustration of (a) training 27 emotion-specific prompts to stimulate the pre-trained large language model (LLM) for emotion inference tasks on the training set. Illustration of (b) obtaining and validating the LLM-based knowledge representation by searchlight representational similarity analysis (RSA; Kriegeskorte et al., 2008) between every artificial neuron's activation in response to 27 prompts and the human ratings on 14 conceptual attributes. According to the rank of Kendall's tau values, the top $N$ neurons most relevant to 14 conceptual attributes were (c) targeted and (d) manipulated during emotion inference tasks on the test set. The example of $N$ = 4,000 was shown in (c). See details in Method and full results of different $N$ in Figs. S5.

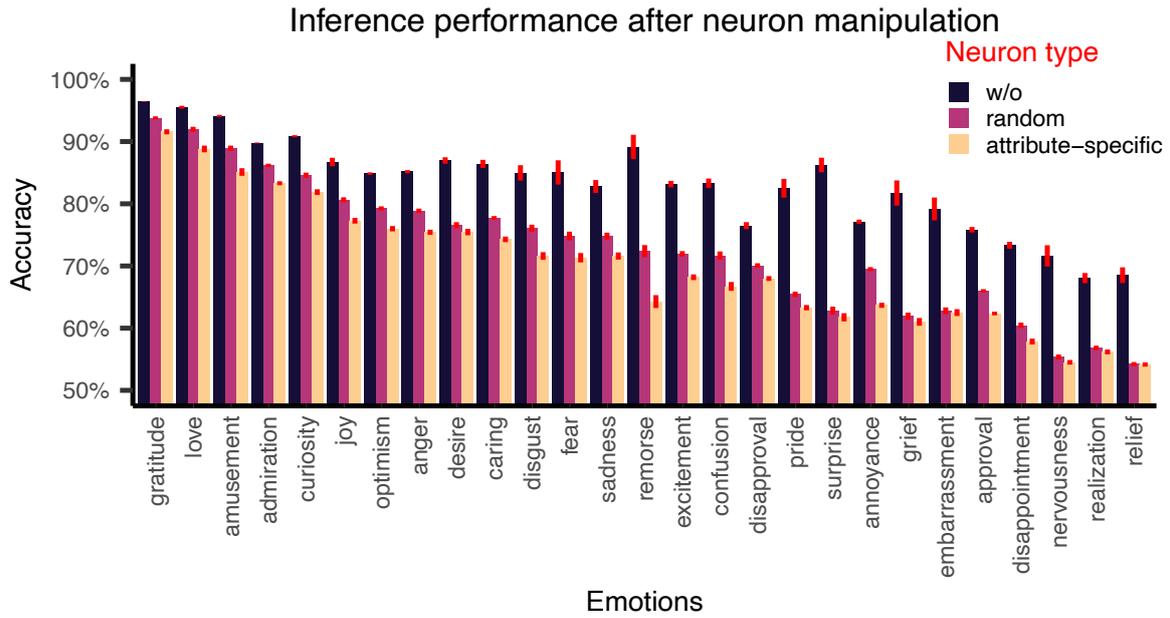

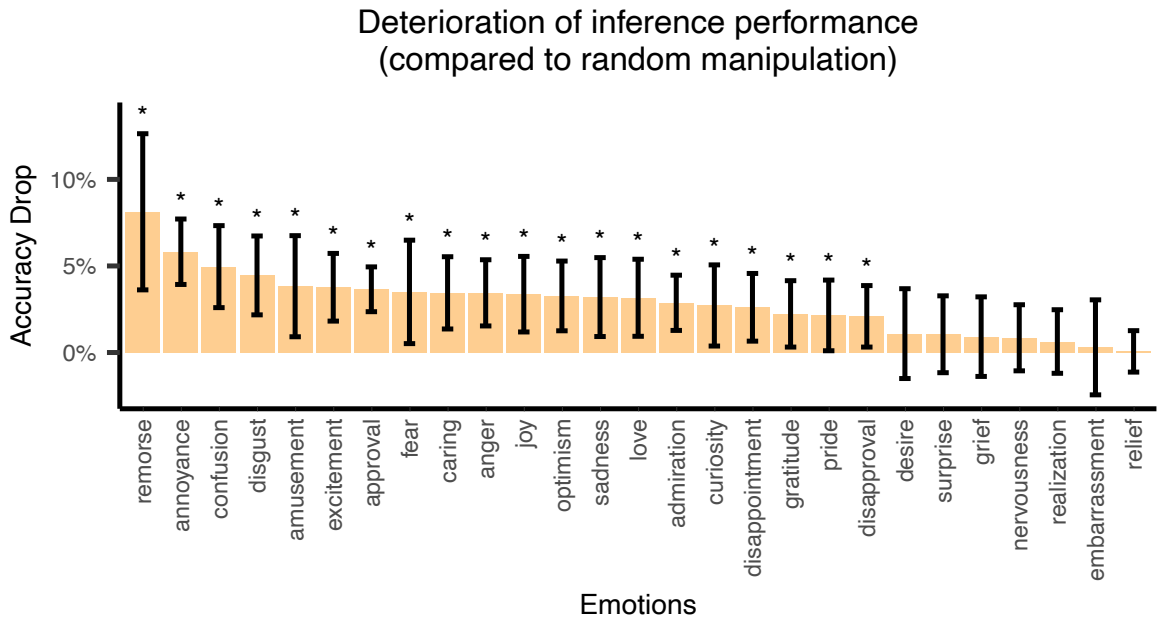

**Figure 3. Results of Manipulating the Language-Specific Representations of Emotion-Concept Knowledge during Emotion Inference Tasks.**

The inference accuracy was evaluated (a) with manipulating $N$ random neurons or without manipulation. Error bars indicate standard errors. The causal contribution of emotion-concept knowledge was indicated by (b) the accuracy drops due to manipulated neuron type (attribute-specific vs. random). Asterisks indicate $p < .05$ with false discovery rate (FDR; Benjamini & Yekutieli, 2001) corrected across all inference tasks and numbers of manipulated neurons. Error bars indicate 95% CIs. The example of $N = 4{,}000$ was shown in (a and b). See details in Method and full results of different $N$ in Figs. S6-7.

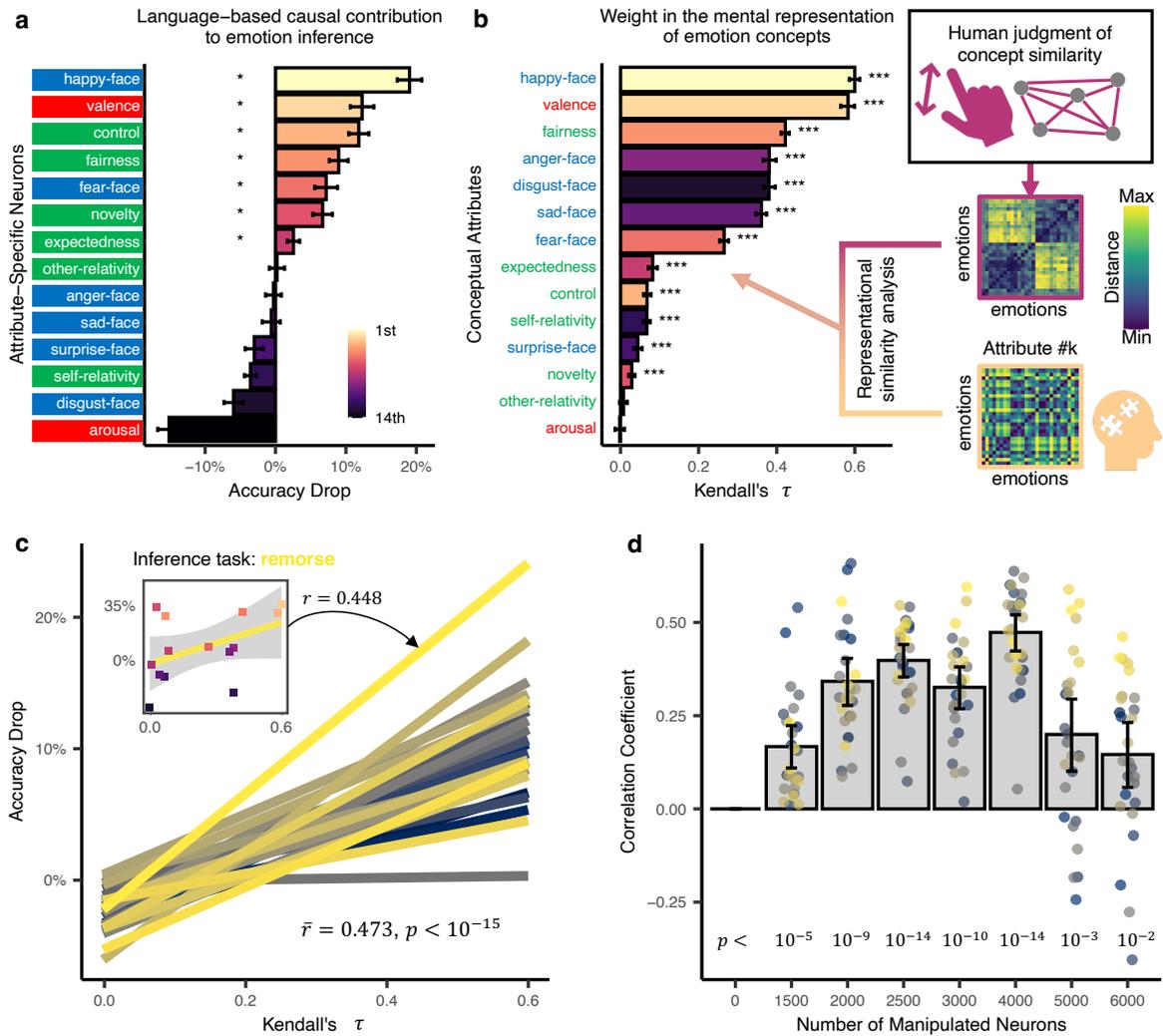

**Figure 4. Comparison between the Language-Based Knowledge Contribution to Emotion Inference and Knowledge Weight in the Human Mental Representation of Emotion Concepts.**

(a) The language-based knowledge contribution to emotion inference. (b) Knowledge weight in the human mental representation of emotion concepts. The asterisks in the former indicate $p < .05$ with FDR corrected across all conceptual attributes and numbers of manipulated neurons. Here, the example of manipulating 4,000 neurons was shown (see Fig. S10 for full results). The latter was estimated by RSA, and the human mental representation of emotion-concepts ($N = 61$) was obtained by the concept similarity judgment experiment. The asterisks indicate FDR corrected bootstrap-based $p < .001$ across all conceptual attributes. See Method for details. For both (a and b), colors of attribute labels indicate "core affects (red)", "prototypical expressions (blue)", and "antecedent appraisals (green)". The conceptual attributes' weights in human mental representation were then correlated to language-based contributions on (subplot in c) arbitrary emotion inference task when (c) manipulating the top 4,000 attribute-specific neurons and (d) manipulating a different number of neurons. Lines in (c) are fitted by linear regression. All error bars indicate 95% CIs.

**Supplementary Information**

**Language-Specific Representation of Emotion-Concept Knowledge Causally Supports Emotion Inference**

Ming Li,[1,2,#] Yusheng Su,[3,#] Hsiu-Yuan Huang,[4] Jiali Cheng,[5] Xin Hu,[1,2,6] Xinmiao Zhang,[1,2] Huadong Wang,[3] Yujia Qin,[3] Xiaozhi Wang,[3] Kristen A. Lindquist,[7] Zhiyuan Liu,[3,**] Dan Zhang[1,2,*]

[1]Department of Psychology, Tsinghua University

[2]Tsinghua Laboratory of Brain and Intelligence, Tsinghua University

[3]Department of Computer Science and Technology, Tsinghua University

[4]School of Computer and Communication Engineering, University of Science and Technology Beijing

[5]Miner School of Computer and Information Sciences, University of Massachusetts Lowell

[6]Department of Psychiatry, University of Pittsburgh

[7]Department of Psychology and Neuroscience, University of North Carolina

**Contact Information**

Correspondence concerning this article should be addressed to Dan Zhang (dzhang@tsinghua.edu.cn) and Zhiyuan Liu (liuzy@tsinghua.edu.cn).

**This PDF file includes:**

    Supplemental Methods

    Figures S1 to S10

    Tables S1 to S2

    SI References

[#]These authors contributed equally.

**Supplemental Methods**

**Training Emotion-Specific Task Prompts**

Formally, $\mathcal{M}$ was a LLM model, RoBERTa (Liu et al., 2019). Given an input text with $n$ tokens $X = \{w_1, w_2, \cdots, w_n\}$, RoBERTa first converted them into input embeddings $X_e \in R^{n \times d}$, where $d$ was the dimension of the embedding space. We pre-pended $l$ randomly initialized trainable tokens $P_e \in R^{l \times d}$ before the input matrix $X_e$, and formed the modified input embeddings $[P_e, X_e] \in R^{(l+n) \times d}$. A special [MASK] token was additionally pre-pended before the prompts, which would output the probability of label tokens. The objective (O) was to maximize the likelihood of the desired output $y$:

$$O = P_{\mathcal{M}}([MASK] = y \mid [P_e, X_e]).$$

During the prompt tuning, we only optimized the trainable tokens ($P_e$) while freezing the whole parameters of a RoBERTa ($\mathcal{M}$) to maximize the above objective.

To obtain the corresponding prompt of each emotion on RoBERTa, we re-framed the 27-class emotion dataset of GoEmotions (Demszky et al., 2020) into 27 emotion inference tasks. For instance, for the emotion "remorse", if a text belonged to the category "remorse", then we re-labeled the text with $y$ = "yes"; otherwise, $y$ = "no". In this way, we obtained the new training data for each emotion. During training, we set the prompt length to $l = 100$ and the prompt dimension to $d = 768$. After conducting prompt tuning individually for each emotion inference task, we obtained all prompts $\{P_e^c \in R^{100 \times 768} \mid c \in \mathcal{C}\}$, where $\mathcal{C}$ was the set of 27 discrete emotions.

In order to avoid statistical bias, for each emotion inference task, we trained prompts 12 times with 12 random seeds; all of these 12 prompts have been evaluated on the test set, respectively.

**Collecting LLM-Based Knowledge Representation**

Previous works (Dai et al., 2022; Geva et al., 2021) have indicated that the values of the artificial neurons in the feed-forward layers FFN(·) of a LLM (Vaswani et al., 2017), RoBERTa, correspond to specific model behaviors. Some studies (Su et al., 2022; Wang et al., 2022) have taken a further step to utilize trained prompts to stimulate RoBERTa and found that the prompts of similar tasks would have similar values of the artificial neurons. In this sense, we hypothesized that the neuron activation values could represent the task-specific knowledge (i.e., emotion-concept knowledge), which could facilitate us to manipulate specific neurons for the purpose of manipulating specific emotion-concept knowledge.

In our setting, the values of artificial neurons $s$ were the values of hidden states between the FFN layer in a Transformer. Specifically, we could denote the FFN layer as:

$$\text{FFN}(x) = \text{GELU}(xW_1^\top + b_1)W_2 + b_2,$$

where $x \in R^d$ was the input embedding, $W_1, W_2 \in R^{d_m \times d}$ were trainable matrices, and $b_1, b_2$ were bias vectors. The value of artificial neurons was $v = xW_1^\top + b_1$.

For each task, we input the sequence, $\{[MASK], P, <s>\}$, into RoBERTa, where $P$ was the emotion-specific prompt, $<s>$ was the special token indicating the start of an input sentence. Finally, we stacked the values of artificial neurons in all FFN layers of RoBERTa to get the overall neuron activation values AS($P$) (https://github.com/thunlp/Prompt-Transferability) for each emotion inference task:

$$\text{AS}(P) = [v_1; v_2; \ldots; v_L],$$

where $L = 36,864$ was the total number of artificial neurons.

**Conceptual Attributes Rating Experiments**

*Core Affects*

Thirty participants (15 females, mean age = 33 years) were recruited to rate 27 emotion concepts' core affects (arousal and valence) directly. The emotions were presented randomly for each participant, followed by their literal definition (consistent with the GoEmotions dataset; Demszky et al., 2020) and a nine-point Likert scale for both attributes. There was text instruction above each rating scale, "To what extent does [$EMOTION$] make you feel... (Valence: 1=very unpleasant, 5=neutral, 9=very pleasant; Arousal: 1=very calming, 9=very arousing)".

*Prototypical Expressions*

Another thirty participants (17 females, mean age = 31 years) were recruited to rate the similarities between 6 prototypical expressions and 27 emotion concepts. A nine-point Likert scale was presented to participants, and each participant's order of emotions (with literal definition) and faces were randomized. There was text and image instruction above each rating scale, "To what extent is $[EMOTION]$ consistent with the physiological responses shown in the figures: (1=very inconsistent, 5=neutral, 9=very consistent)". The images we used to indicate six prototypical expressions are twelve averaged faces (one male and one female for each prototypical expression) from the AKDEF stimulus set (https://kdef.se/index.html; Lundqvist & Flykt, 1998).

### *Antecedent Appraisals*

Two hundred ninety-nine independent participants (148 female, mean age = 37 years) were recruited to recall an event that caused them to feel one of the 27 emotions (randomly assigned) and rate 38 items on the event. In the recall phase, we instructed participants to remember and write down a situation (at least 100 words) in which they felt the given emotion (with literal definition) and then identify the specific event (up to 50 words) in the situation that directly caused that emotion. This procedure avoided involving multiple events, cognitions, and emotions in a single recall (Roseman et al., 1990). We instructed participants in the next phase to rate 38 items for that specific event in random order. All those items were summarized by a previous study (Skerry et al., 2015), covering most factors from the appraisal theories of emotion. Before the next processing step, we kept six factors in the 299 events times 38 items matrix as appraisal attributes (see Fig. S2 for further details).

**Figures S1 to S10**

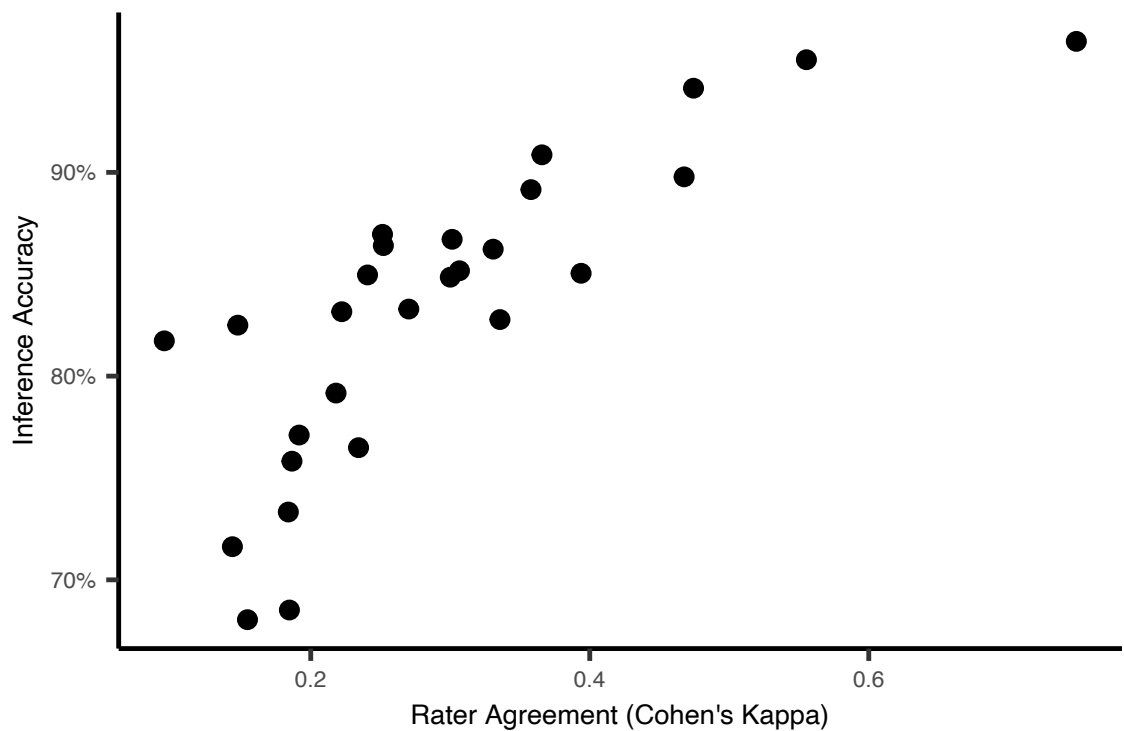

**<u>Figure S1. The LLM's Inference Accuracy for 27 Emotions Is Significantly Related to the Rater Agreement for Annotating Corresponding Emotions.</u>**

The rater agreement for annotating corresponding emotions was estimated by Cohen's Kappa (Cohen, 1960). Pearson's *r*(25) = .797, *p* < .001. As the rater agreement reflects the extent to which people share the conceptualization of specific emotions, the positive correlation suggests that what the LLM has learned in the large-scale language corpus is a shared understanding among individuals.

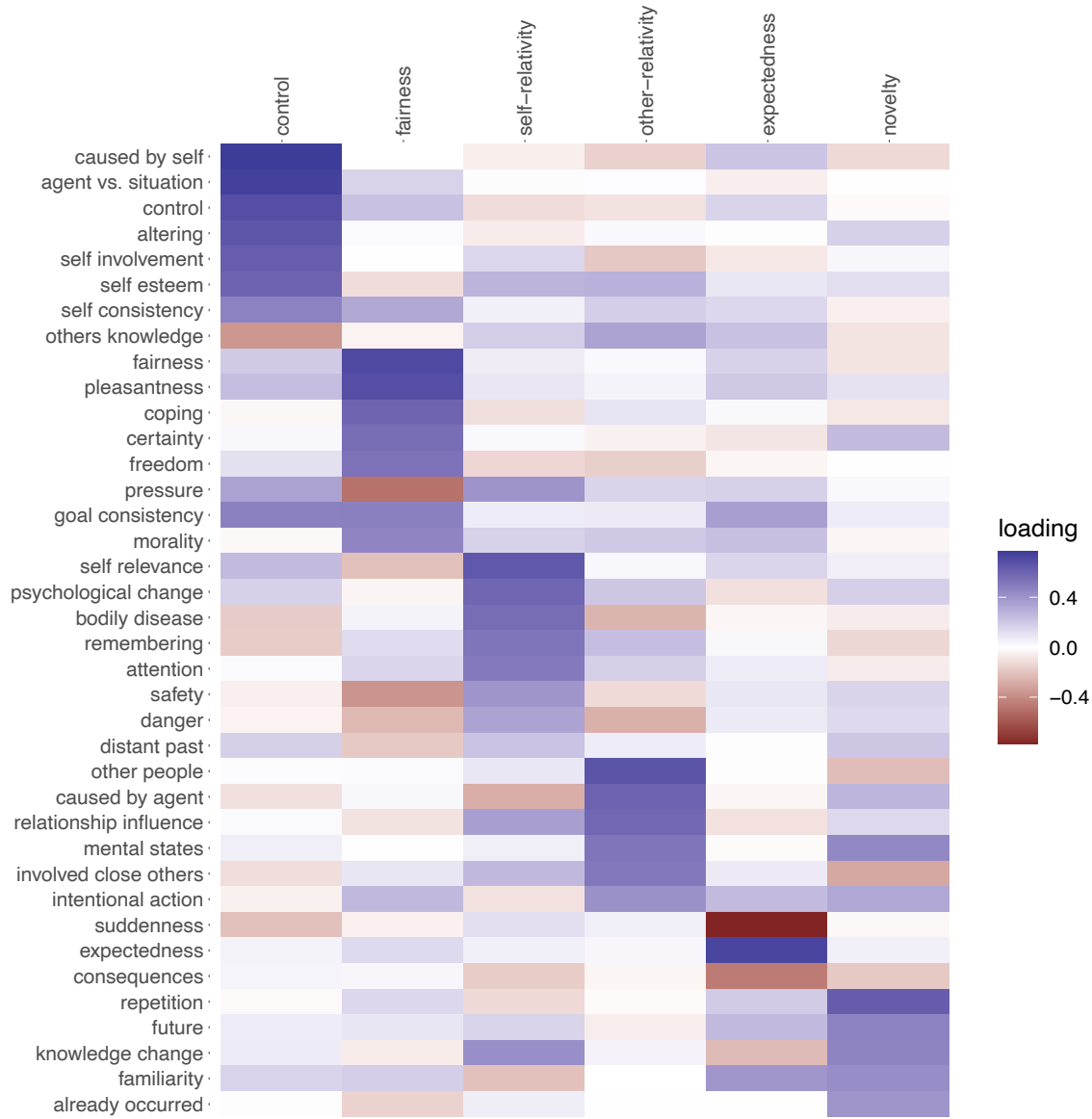

**Figure S2. Six Antecedent Appraisal Attributes of Emotion Were Extracted from the Raw Score Matrix (299 Participants by 38 Items) Obtained from the Conceptual Attribute Rating Experiment.**

Each participant was instructed to recall an event that directly caused them to feel a given emotion (randomly assigned) and rate 38 items on the event. All those items were summarized by Skerry et al. (2015), covering most factors from the appraisal theories of emotion. Each of the 27 emotions was rated by at least 11 participants. The number of extracted factors was determined based on parallel analysis. The extraction method is Principal Component Analysis. The rotation method is Varimax (with Kaiser Normalization). The color indicates the loadings (correlation coefficients) between appraisal attributes (columns) and event items (rows).

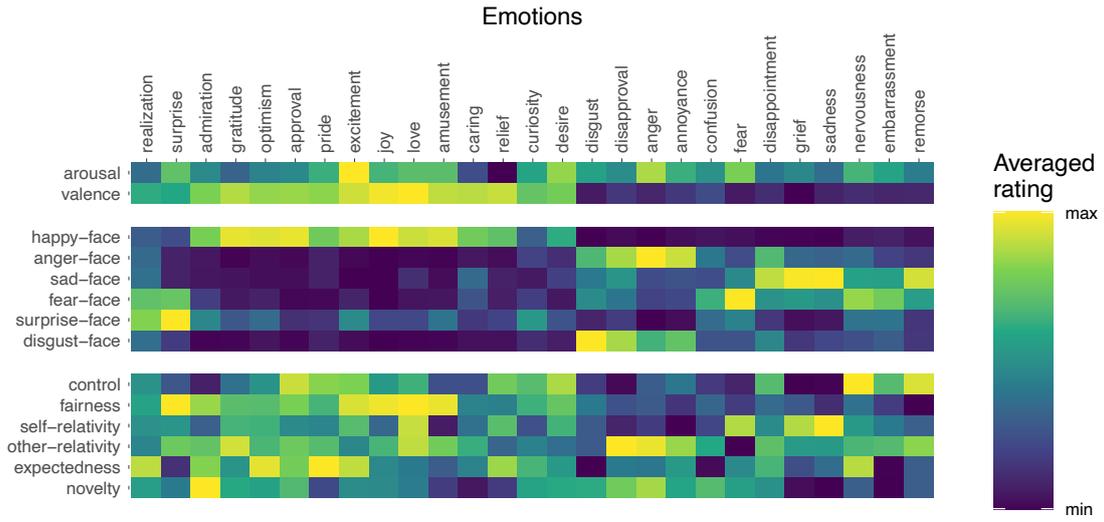

**Figure S3. Averaged Human Ratings for 14 Conceptual Attributes (Two Core Affects, Six Prototypical Expressions, and Six Antecedent Appraisals) of 27 Emotions.**

Color bar indicates the standardized rating scores.

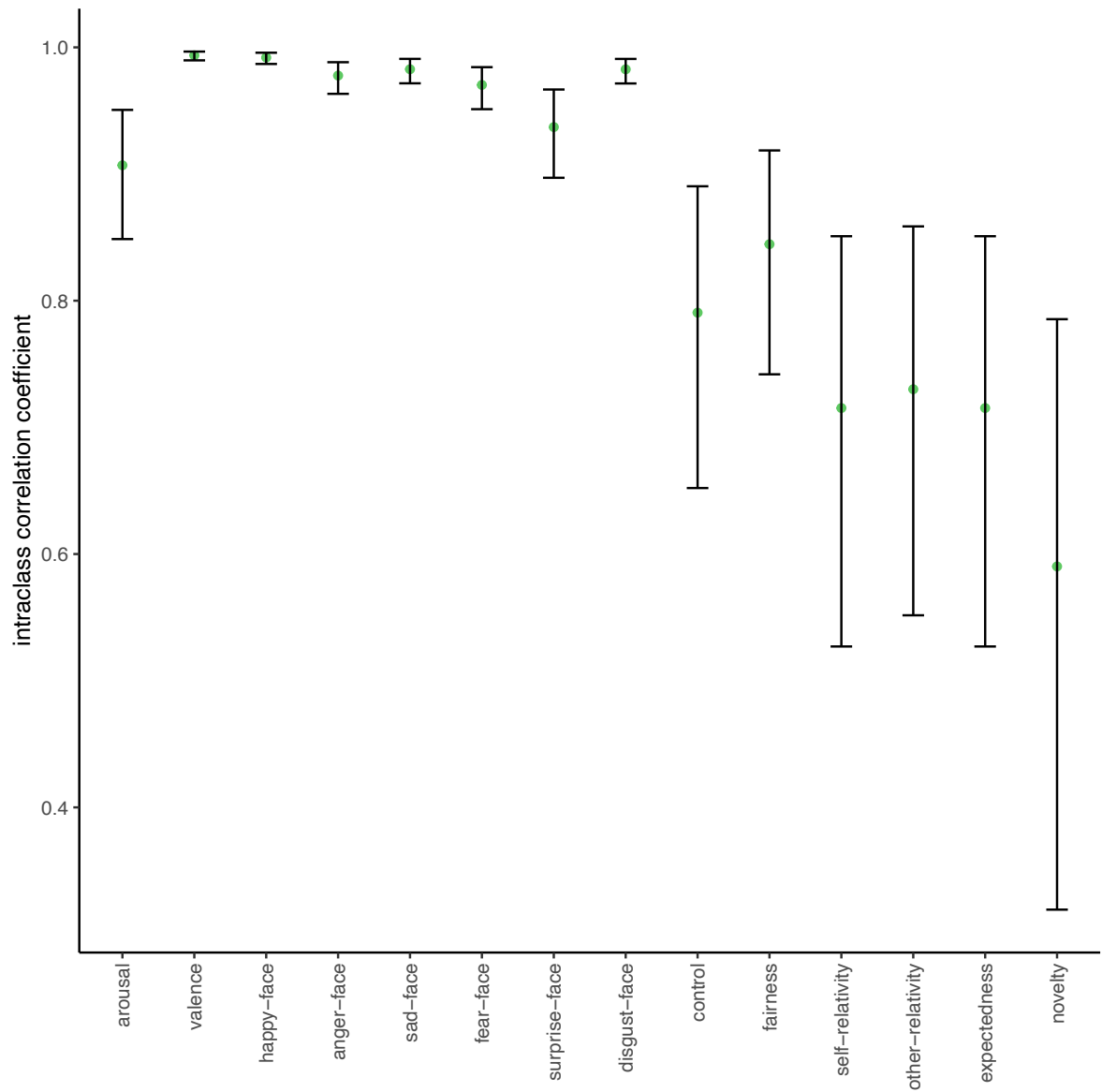

**Figure S4. The Reliabilities of 14 Conceptual Attributes Were Determined by the Intra-Class Correlation Coefficient (ICC; Koo & Li, 2016).**

According to the way the three types of conceptual attributes were obtained, their reliabilities were calculated as ICC(2,k) for core affects, ICC(2,k) for prototypical expressions, and ICC(1,k) for antecedent appraisals, respectively. All the ICCs were significant at the level = .001 with FDR corrected. Error bars indicate the 95% CIs.

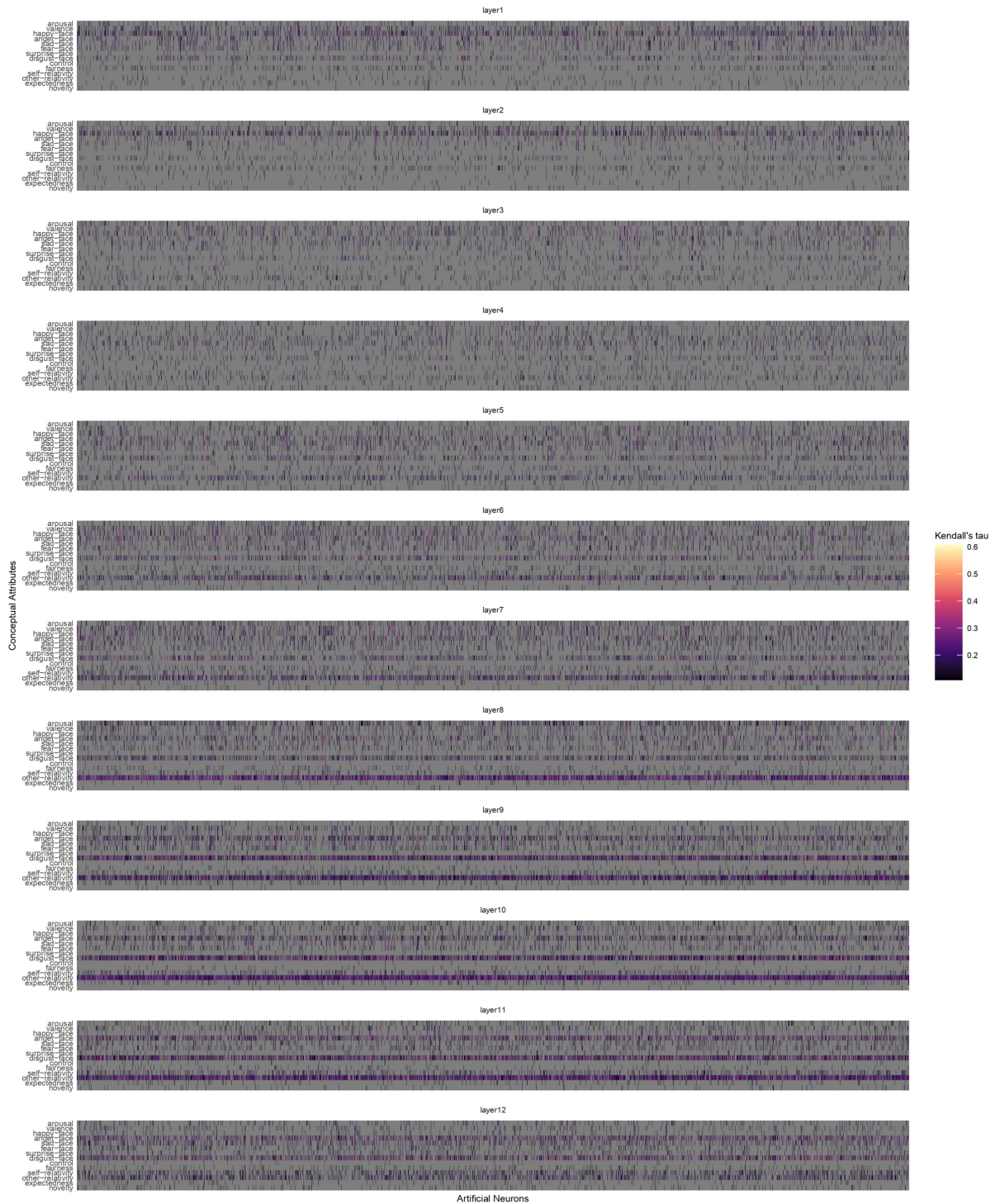

**Figure S5. Representational Similarity Analysis (RSA) Results.**

Representational similarity analysis (RSA; Kriegeskorte et al., 2008) results for relating representational dissimilarity matrices (RDMs) of every single artificial neuron's activation and the RDMs of 14 conceptual attributes of emotion. Each of the 14 conceptual attributes was significantly related to a subset of artificial neurons with $p < .01$ across all comparisons with false discovery rate (FDR; Benjamini & Yekutieli, 2001) corrected for one-tailed sign-rank test. Color bar indicates the significant tau values. Gray color indicates no significance.

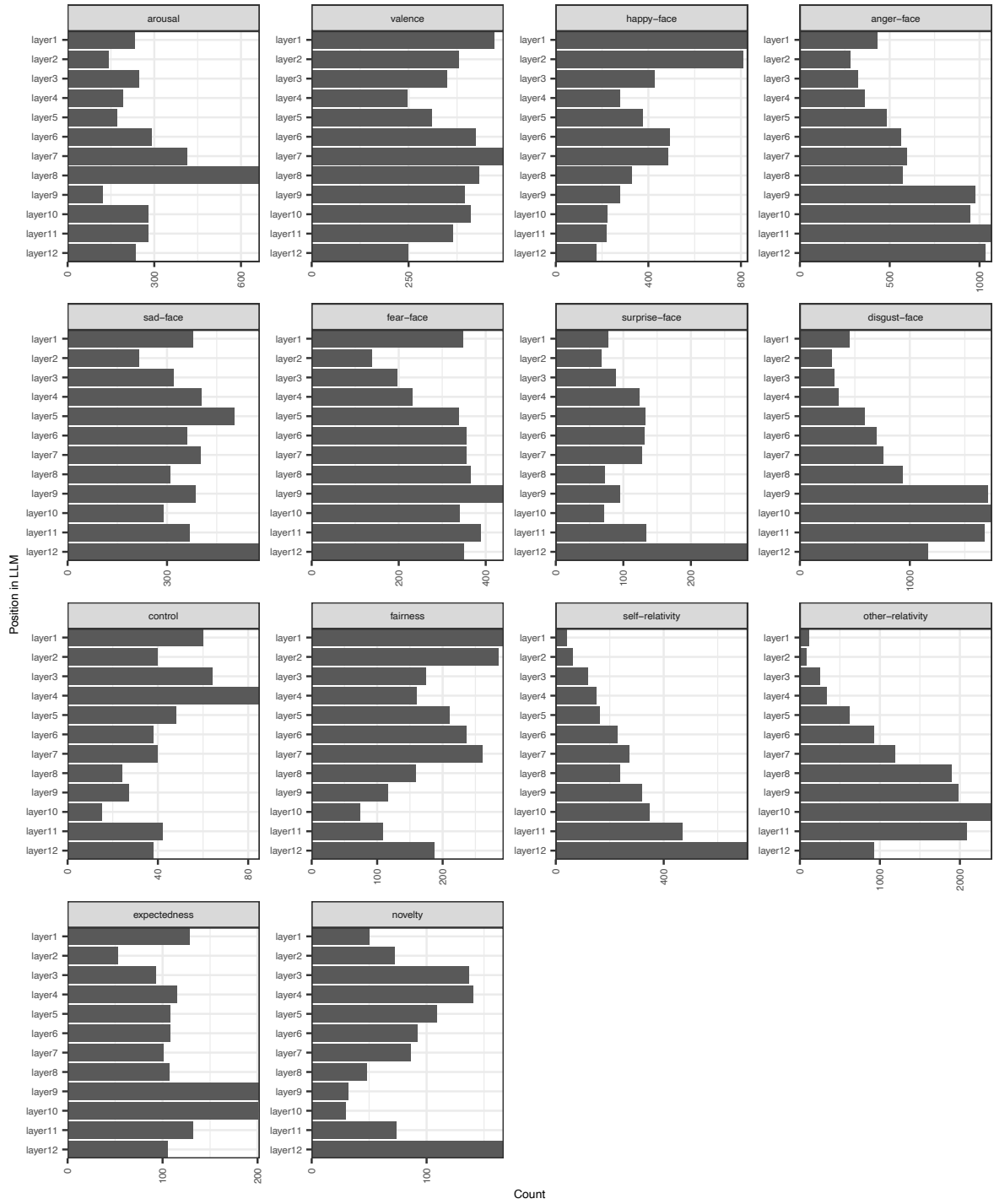

**Figure S6. The Count of the Significant Relevant Artificial Neurons for 14 Conceptual Attributes in Each Layer of the LLM.**

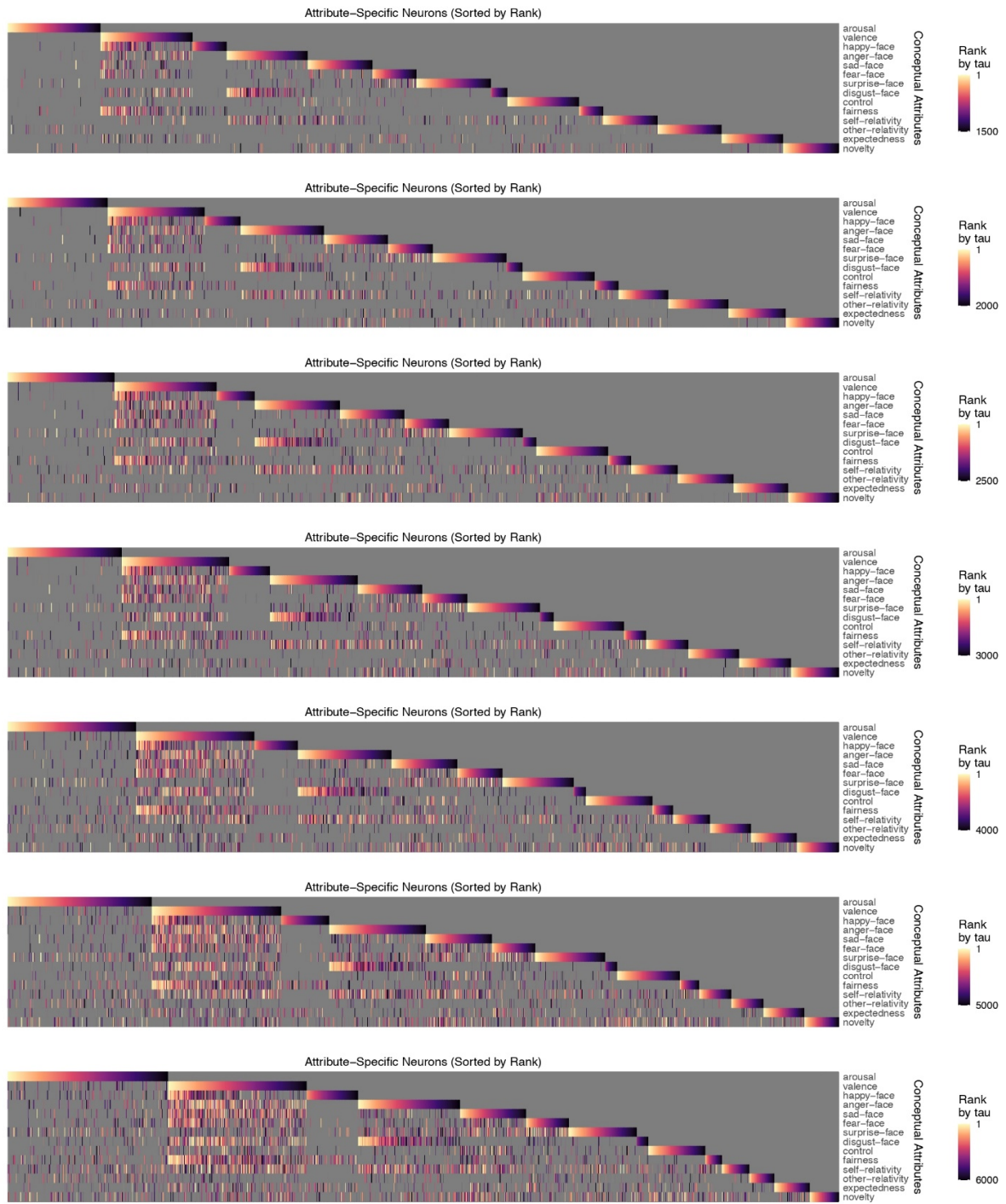

**Figure S7. Relative Correspondence between Artificial Neurons and Conceptual Attributes of Emotions.**

According to the rank of Kendall's tau values between neuron RDMs and each attribute RDM (color bar), the top *N* neurons most relevant to 14 conceptual attributes tend to be distinct. From top to bottom, *N* = 1500, 2000, 2500, 3000, 4000, 5000, and 6000.

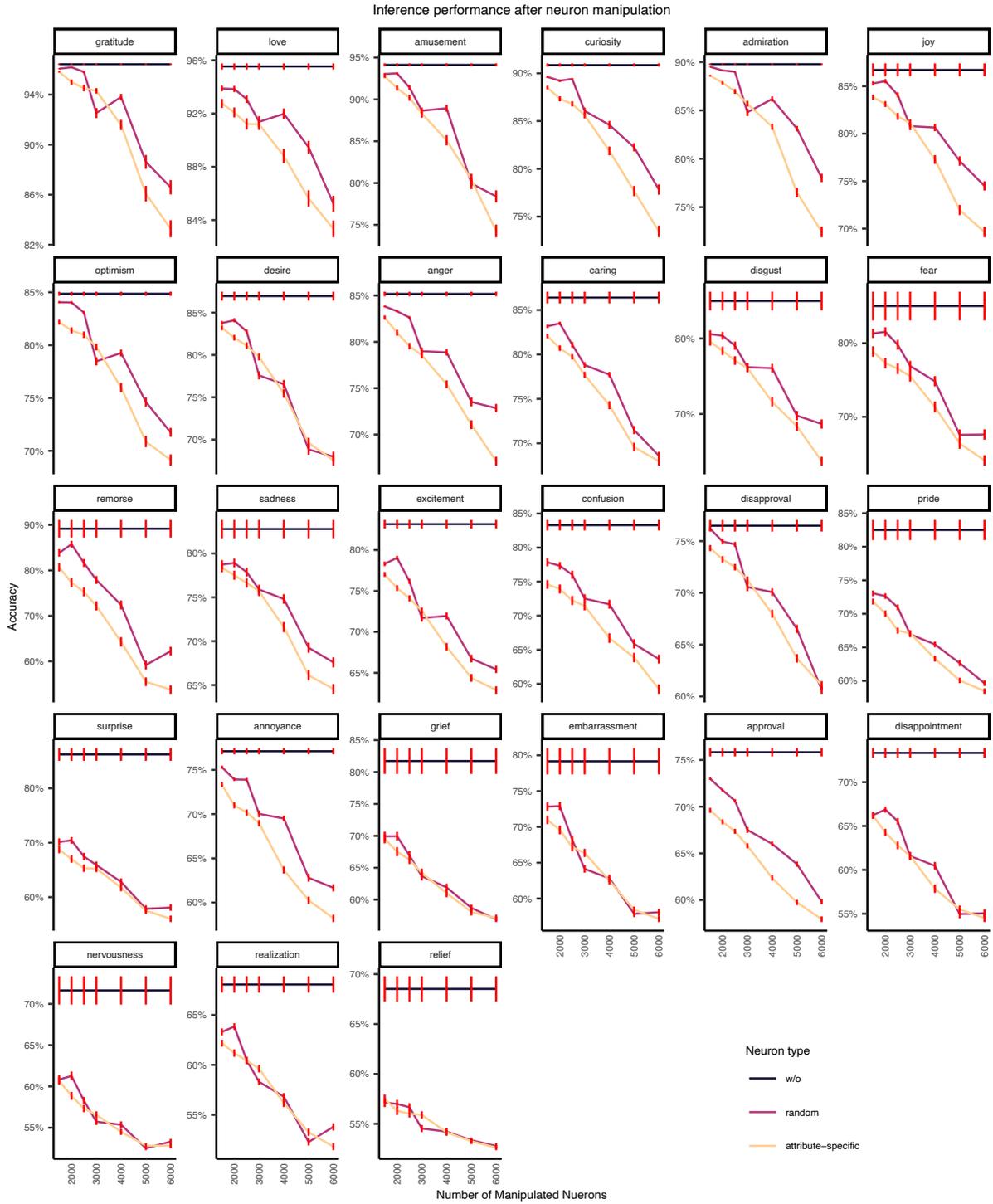

**Figure S8. The Accuracy of 27 Emotion Inference Tasks on the Test Set with Each Panel Corresponding to An Emotion.**

By targeting and manipulating the top *N* (x-axis) attribute-specific neurons during different emotion inference tasks, the inference performance (light orange lines) deteriorated compared to without manipulation (black lines). We also randomly manipulated *N* neurons (pink lines) to exclude the influence of manipulation *per se*. Error bars indicate standard error.

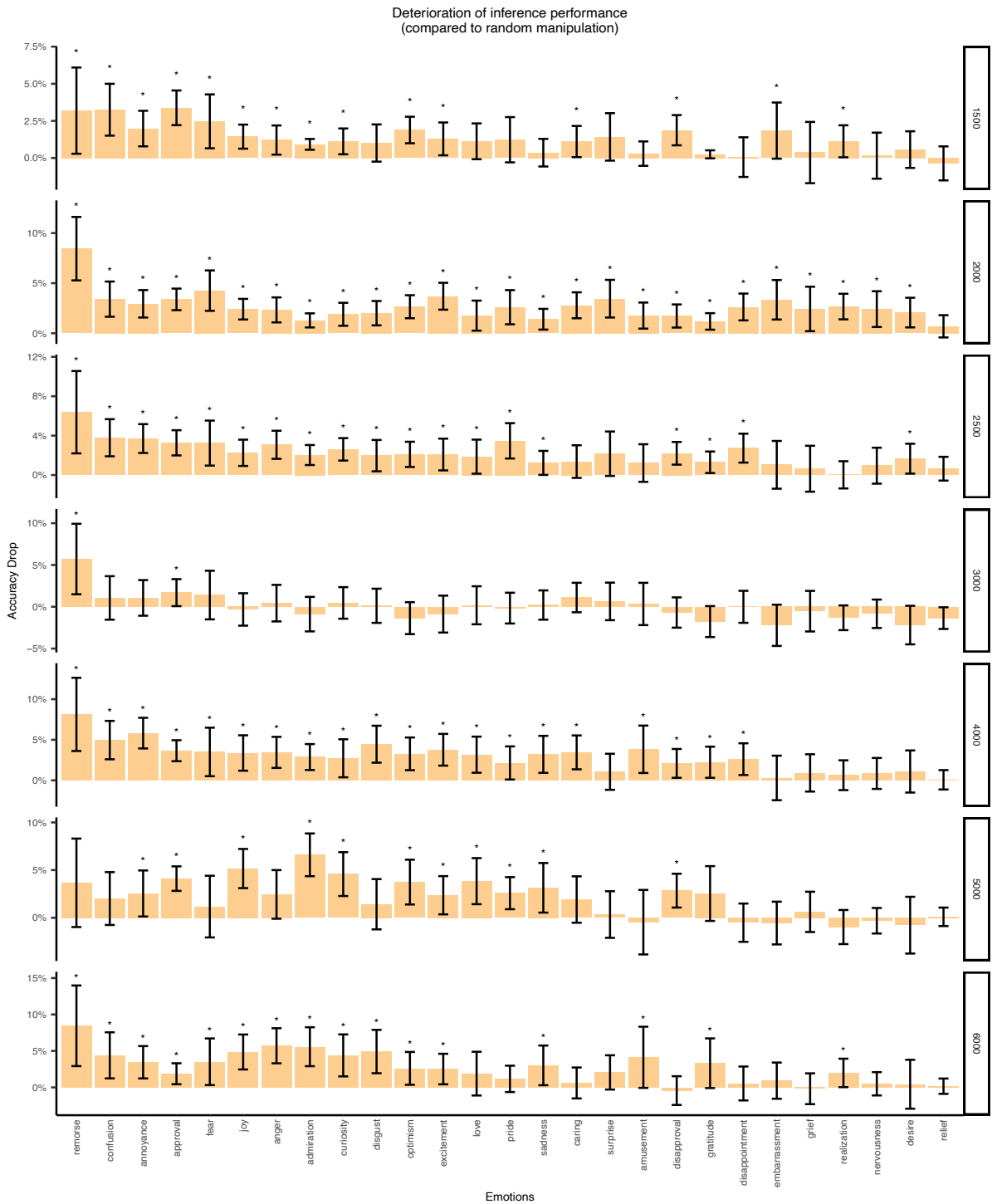

**Figure S9. Causal Contribution of Emotion-Concept Knowledge to 27 Emotion Inference Tasks.**

The causal contributions when manipulating the top *N* attribute-specific neurons are indicated by the difference in accuracy compared to random manipulation. Asterisks indicate FDR corrected *p* < .05 across all inference tasks and numbers of manipulated neurons, one-tailed paired *t*-test for random seeds and conceptual attributes. Error bars indicate 95% CIs. From top to bottom, *N* = 1500, 2000, 2500, 3000, 4000, 5000, and 6000.

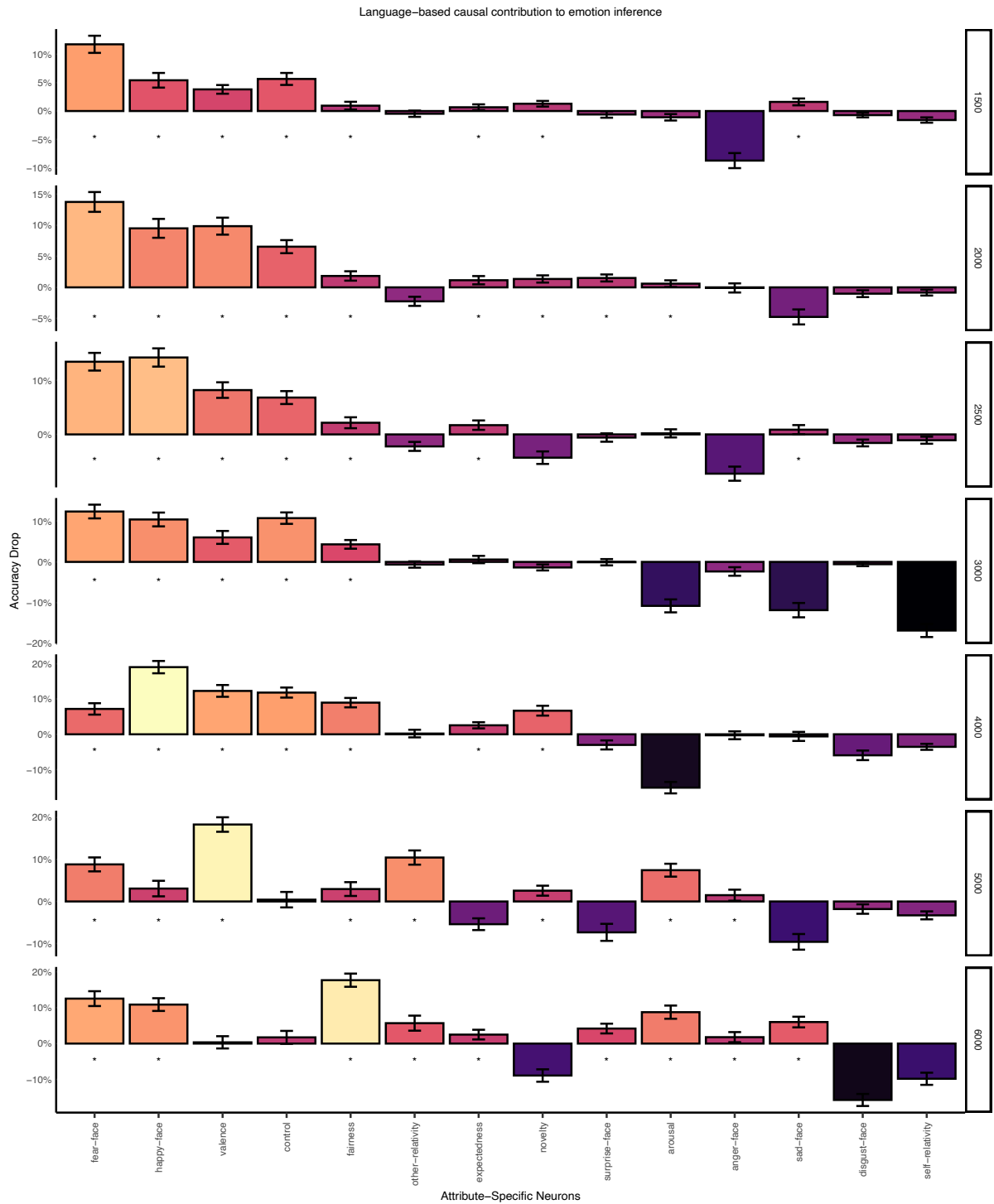

**Figure S10. Causal Contribution from 14 Conceptual Attributes of Emotions.**

The language-based causal contributions of 14 conceptual attributes to emotion inference tasks were indicated as the accuracy drop after manipulating the top *N* attribute-specific artificial neurons (compared to randomly manipulating the same number of neurons). Asterisks indicate FDR corrected $p < .05$ across 14 conceptual attributes and seven numbers of manipulated neurons, one-tailed paired *t*-test for random seeds and inference tasks. Error bars indicate 95% CIs. From top to bottom, N = 1500, 2000, 2500, 3000, 4000, 5000, and 6000.

**Tables S1 to S2**

**Table S1**

*The LLM's Inference Accuracies and Human Rater Agreements on 27 Emotions*

| Emotion | Human Raters' Agreement (Cohen's kappa) | LLM's Inference Accuracy (%) | |
|---|---|---|---|
| | | M | SD |
| admiration | 0.468 | 89.8 | 0.5 |
| amusement | 0.474 | 94.1 | 1.2 |
| anger | 0.307 | 85.2 | 1.7 |
| annoyance | 0.192 | 77.1 | 2.5 |
| approval | 0.187 | 75.8 | 3.1 |
| caring | 0.252 | 86.4 | 4.6 |
| confusion | 0.270 | 83.3 | 5.4 |
| curiosity | 0.366 | 90.9 | 1.2 |
| desire | 0.251 | 87.0 | 3.6 |
| disappointment | 0.184 | 73.3 | 3.8 |
| disapproval | 0.234 | 76.5 | 3.9 |
| disgust | 0.241 | 85.0 | 8.6 |
| embarrassment | 0.218 | 79.2 | 12.7 |
| excitement | 0.222 | 83.2 | 3.6 |
| fear | 0.394 | 85.0 | 13.5 |
| gratitude | 0.749 | 96.4 | 0.4 |
| grief | 0.095 | 81.7 | 14.0 |
| joy | 0.301 | 86.7 | 4.7 |
| love | 0.555 | 95.5 | 1.6 |
| nervousness | 0.144 | 71.6 | 11.9 |
| optimism | 0.300 | 84.9 | 1.4 |
| pride | 0.148 | 82.5 | 10.5 |
| realization | 0.155 | 68.0 | 5.7 |
| relief | 0.185 | 68.5 | 8.7 |
| remorse | 0.358 | 89.2 | 13.5 |
| sadness | 0.336 | 82.8 | 7.2 |
| surprise | 0.331 | 86.2 | 8.1 |

*Note.* The rater agreements were estimated via Cohen's Kappa (Cohen, 1960). The inference accuracies were calculated over 12 different random seeds by evaluating the trained prompts on the test set.

**Table S2**

*Heterogeneity Degree in Conceptual Attributes' Causal Contribution to Various Emotion Inference Tasks*

| Number of Manipulated Neurons | Conceptual Attribute | Dip statistic | *p*-value | N |
|---|---|---|---|---|
| 1500 | arousal | 0.066 | 0.468 | 27 |
| 1500 | valence | 0.042 | 0.988 | 27 |
| 1500 | happy | 0.062 | 0.561 | 27 |
| 1500 | anger | 0.047 | 0.944 | 27 |
| 1500 | sad | 0.055 | 0.768 | 27 |
| 1500 | fear | 0.065 | 0.484 | 27 |
| 1500 | surprise | 0.047 | 0.937 | 27 |
| 1500 | disgust | 0.070 | 0.359 | 27 |
| 1500 | control | 0.073 | 0.283 | 27 |
| 1500 | fairness | 0.045 | 0.964 | 27 |
| 1500 | self-related | 0.035 | 0.995 | 27 |
| 1500 | other-related | 0.052 | 0.856 | 27 |
| 1500 | expectedness | 0.043 | 0.979 | 27 |
| 1500 | non-novelty | 0.057 | 0.713 | 27 |
| 2000 | arousal | 0.064 | 0.496 | 27 |
| 2000 | valence | 0.060 | 0.637 | 27 |
| 2000 | happy | 0.062 | 0.578 | 27 |
| 2000 | anger | 0.038 | 0.993 | 27 |
| 2000 | sad | 0.079 | 0.175 | 27 |
| 2000 | fear | 0.059 | 0.666 | 27 |
| 2000 | surprise | 0.054 | 0.799 | 27 |
| 2000 | disgust | 0.049 | 0.906 | 27 |
| 2000 | control | 0.069 | 0.386 | 27 |
| 2000 | fairness | 0.066 | 0.444 | 27 |
| 2000 | self-related | 0.044 | 0.974 | 27 |
| 2000 | other-related | 0.052 | 0.849 | 27 |
| 2000 | expectedness | 0.066 | 0.453 | 27 |
| 2000 | non-novelty | 0.037 | 0.994 | 27 |
| 2500 | arousal | 0.050 | 0.895 | 27 |
| 2500 | valence | 0.063 | 0.531 | 27 |
| 2500 | happy | 0.058 | 0.686 | 27 |
| 2500 | anger | 0.048 | 0.924 | 27 |
| 2500 | sad | 0.043 | 0.982 | 27 |
| 2500 | fear | 0.057 | 0.716 | 27 |
| 2500 | surprise | 0.060 | 0.629 | 27 |
| 2500 | disgust | 0.070 | 0.342 | 27 |
| 2500 | control | 0.057 | 0.709 | 27 |
| 2500 | fairness | 0.054 | 0.798 | 27 |
| 2500 | self-related | 0.052 | 0.843 | 27 |
| 2500 | other-related | 0.049 | 0.911 | 27 |
| 2500 | expectedness | 0.070 | 0.359 | 27 |
| 2500 | non-novelty | 0.049 | 0.915 | 27 |

| Number of Manipulated Neurons | Conceptual Attribute | Dip statistic | p-value | N |
|---|---|---|---|---|
| 3000 | arousal | 0.047 | 0.948 | 27 |
| 3000 | valence | 0.064 | 0.519 | 27 |
| 3000 | happy | 0.073 | 0.276 | 27 |
| 3000 | anger | 0.061 | 0.612 | 27 |
| 3000 | sad | 0.044 | 0.976 | 27 |
| 3000 | fear | 0.065 | 0.493 | 27 |
| 3000 | surprise | 0.055 | 0.779 | 27 |
| 3000 | disgust | 0.081 | 0.153 | 27 |
| 3000 | control | 0.051 | 0.881 | 27 |
| 3000 | fairness | 0.042 | 0.987 | 27 |
| 3000 | self-related | 0.055 | 0.774 | 27 |
| 3000 | other-related | 0.049 | 0.908 | 27 |
| 3000 | expectedness | 0.056 | 0.737 | 27 |
| 3000 | non-novelty | 0.061 | 0.613 | 27 |
| 4000 | arousal | 0.070 | 0.346 | 27 |
| 4000 | valence | 0.048 | 0.922 | 27 |
| 4000 | happy | 0.083 | 0.127 | 27 |
| 4000 | anger | 0.053 | 0.831 | 27 |
| 4000 | sad | 0.047 | 0.949 | 27 |
| 4000 | fear | 0.056 | 0.758 | 27 |
| 4000 | surprise | 0.079 | 0.177 | 27 |
| 4000 | disgust | 0.054 | 0.811 | 27 |
| 4000 | control | 0.040 | 0.991 | 27 |
| 4000 | fairness | 0.057 | 0.708 | 27 |
| 4000 | self-related | 0.051 | 0.863 | 27 |
| 4000 | other-related | 0.038 | 0.993 | 27 |
| 4000 | expectedness | 0.057 | 0.733 | 27 |
| 4000 | non-novelty | 0.051 | 0.872 | 27 |
| 5000 | arousal | 0.050 | 0.895 | 27 |
| 5000 | valence | 0.049 | 0.909 | 27 |
| 5000 | happy | 0.055 | 0.790 | 27 |
| 5000 | anger | 0.045 | 0.960 | 27 |
| 5000 | sad | 0.058 | 0.705 | 27 |
| 5000 | fear | 0.049 | 0.916 | 27 |
| 5000 | surprise | 0.065 | 0.480 | 27 |
| 5000 | disgust | 0.047 | 0.949 | 27 |
| 5000 | control | 0.049 | 0.908 | 27 |
| 5000 | fairness | 0.059 | 0.661 | 27 |
| 5000 | self-related | 0.066 | 0.466 | 27 |
| 5000 | other-related | 0.048 | 0.921 | 27 |
| 5000 | expectedness | 0.054 | 0.796 | 27 |
| 5000 | non-novelty | 0.054 | 0.796 | 27 |
| 6000 | arousal | 0.057 | 0.730 | 27 |
| 6000 | valence | 0.047 | 0.940 | 27 |

| Number of Manipulated Neurons | Conceptual Attribute | Dip statistic | *p*-value | N |
|---|---|---|---|---|
| 6000 | happy | 0.035 | 0.995 | 27 |
| 6000 | anger | 0.052 | 0.844 | 27 |
| 6000 | sad | 0.057 | 0.715 | 27 |
| 6000 | fear | 0.061 | 0.613 | 27 |
| 6000 | surprise | 0.046 | 0.953 | 27 |
| 6000 | disgust | 0.047 | 0.943 | 27 |
| 6000 | control | 0.046 | 0.950 | 27 |
| 6000 | fairness | 0.049 | 0.909 | 27 |
| 6000 | self-related | 0.046 | 0.955 | 27 |
| 6000 | other-related | 0.057 | 0.734 | 27 |
| 6000 | expectedness | 0.046 | 0.954 | 27 |
| 6000 | non-novelty | 0.066 | 0.451 | 27 |

*Note.* The heterogeneity was tested by determining whether the distribution of the given conceptual attribute's causal contribution (accuracy drop compared to random condition) across 27 emotion inference tasks was multimodal (i.e., at least bimodal) using Hartigan's dip statistic (Freeman & Dale, 2013; Gu, M; Lai et al., 1985), with unimodality as the alternative hypothesis.